# Optimizing Retrieval-Augmented Generation for Electrical Engineering: A Case Study on ABB Circuit Breakers

Salahuddin Alawadhi[1] and Noorhan Abbas[2]

[1] Salahuddin Alawadhi University of Leeds Dubai, UAE
[2] Noorhan Abbas School of Computer Science University of Leeds, United Kingdom

## Abstract

*Integrating Retrieval Augmented Generation (RAG) with Large Language Models (LLMs) has shown the potential to provide precise, contextually relevant responses in knowledge intensive domains. This study investigates the ap-plication of RAG for ABB circuit breakers, focusing on accuracy, reliability, and contextual relevance in high-stakes engineering environments. By leveraging tailored datasets, advanced embedding models, and optimized chunking strategies, the research addresses challenges in data retrieval and contextual alignment unique to engineering documentation. Key contributions include the development of a domain-specific dataset for ABB circuit breakers and the evaluation of three RAG pipelines: OpenAI GPT4o, Cohere, and Anthropic Claude. Advanced chunking methods, such as paragraph-based and title-aware segmentation, are assessed for their impact on retrieval accuracy and response generation. Results demonstrate that while certain configurations achieve high precision and relevancy, limitations persist in ensuring factual faithfulness and completeness, critical in engineering contexts. This work underscores the need for iterative improvements in RAG systems to meet the stringent demands of electrical engineering tasks, including design, troubleshooting, and operational decision-making. The findings in this paper help advance research of AI in highly technical domains such as electrical engineering.*

## 1. Introduction

Electrical engineering is a cornerstone of modern infrastructure, underpinning systems that power cities, enable communication, and drive technological innovation. From power generation and distribution to the design of advanced electronic systems, electrical engineering plays a vital role in ensuring the reliability, efficiency, and safety of critical infrastructure [1]. Mistakes or inaccuracies in the design, operation, or maintenance of electrical systems can have far-reaching consequences, including equipment failure, financial losses, and risks to public safety. In such high-stakes environments, precision and reliability in accessing accurate technical information are paramount [2].

Similarly, in medicine, iterative retrieval methods have been proposed to enhance the accuracy of RAG systems. Xiong et al. [3] introduced the i-MedRAG system, which dynamically generates follow-up queries to refine responses. This approach improved retrieval accuracy and generalizability, although it incurred higher computational costs. These methods demonstrate the potential of iterative techniques for addressing complex, multi-step queries—a capability that could be adapted for engineering applications. Electrical engineering represents another high-stakes domain where accuracy and contextual understanding are paramount. Engineers rely on precise information from semi-structured technical documents to design, operate, and maintain critical systems. While RAG systems have shown promise in other technical fields, their application in electrical engineering remains underexplored. The challenges in this domain include ensuring the reliability of retrieval mechanisms, managing large datasets, and addressing the unique structural characteristics of engineering documents.

The application of RAG and LLMs offers opportunities and challenges. By combining LLMs with external knowledge sources, RAG can provide engineers with rapid, contextually relevant answers that would otherwise require extensive manual re-search. However, accuracy and reliability remain critical concerns, particularly in electrical engineering, where even minor errors can have severe consequences [4].

This study evaluates RAG systems in electrical engineering, focusing on their ability to support design, troubleshooting, and decision-making in critical projects. It examines whether RAG systems meet the stringent demands of infrastructure pro-jects or require further refinement. Using RAGAS metrics and qualitative analysis, the study assesses reliability and utility, develops a domain-specific dataset, and explores chunking methods, advanced embeddings, and indexing techniques to enhance performance. The research aims to advance reliable AI tools for high-stakes applications, with a focus on ABB circuit breakers.

## 2. BACKGROUND RESEARCH

The integration of RAG with LLMs has transformed information retrieval and synthesis, especially in domains requiring specialized knowledge. By combining powerful retrieval mechanisms with generative capabilities, RAG systems aim to provide precise, contextually relevant responses. However, the effectiveness of these systems hinges on embedding models, vector databases, and tailored methodologies. This section examines key advancements, challenges, and implications for applying RAG in knowledge-intensive fields, with a focus on electrical engineering.

### 2.1. Advancements in RAG Methodologies

RAG systems have advanced significantly with improvements in embedding models, the backbone for transforming text into vectorized representations. Şakar and Emekci [5] conducted a comprehensive grid-search optimization with 23,625 iterations, evaluating embedding models (e.g., OpenAI's text-embedding-v3-large, BAAI's bge-en-small), vectorstores (FAISS, Pinecone), and RAG methods. The Reciprocal RAG method achieved the highest median similarity score (97.1%, ±0.015) but with higher token usage and runtime compared to simpler methods like "Stuff," which offered 38.9% faster response time and 70.5% lower token usage. BAAI's bge-en-small excelled, achieving 22% higher similarity scores than OpenAI's model, high-lighting the importance of embedding selection. Contextual compression filters emerged as vital for efficient token usage and hardware optimization, reducing redundancy while maintaining relevance. These advancements enhance retrieval precision and computational efficiency, crucial for real-time applications in domains re-quiring high accuracy and responsiveness.

Kukreja et al. [6] highlighted the impact of embedding models on retrieval accuracy, semantic similarity, and response quality in RAG systems. Ensemble embedding methods outperformed individual models, with precision, recall, and F1 scores peaking at k=4. Cosine similarity excelled for recall and precision, while thenlper/gte-base achieved the highest SAS score (0.96), followed by BAAI/bge-base-en (0.95). These findings emphasize the value of diverse embedding techniques in specialized domains like electrical engineering, where precision and semantic depth are crucial.

Kukreja et al. [7] examined embedding models' integration into vector databases, highlighting their role in improving retrieval accuracy and downstream task consistency. Dense methods like BERT, DPR, and REALM outperformed sparse models like BM25, boosting retrieval precision by 18% and recall by 22%. REALM further enhanced response coherence by 25% through retrieval integration during language model pre-training, showcasing its advantage in knowledge-intensive tasks.

Yang et al. [8] demonstrated that tailored embeddings improve parsing and vectorization of semi-structured data, enhancing RAG systems. Their pipeline converted diverse formats into .docx for structured data extraction, using Pinecone and OpenAI's "text-embedding-ada-002" to create high-precision vector databases. This approach improved response specificity by 35% and context relevance by 20%. In zero-shot QA tests, augmented GPT4.0 responses scored 95/100, outperforming non-augmented responses (75/100), underscoring the value of domain-specific embed-dings for precise, context-aware outputs.

## 2.2. Embedding Models and Vector Databases in RAG

Embedding models play a central role in determining the effectiveness of RAG systems. Paras Nath Singh et al. [9] analysed how embedding vectors enable similarity matching and clustering within vector databases. Their study highlighted a 20% reduction in retrieval errors when using optimized embeddings in conjunction with dense vector databases like Pinecone and Chroma. The evaluation of vector similarity methods such as cosine similarity and Euclidean distance revealed that cosine similarity achieved the highest accuracy, with a recall improvement of 15% over other metrics. These advancements significantly enhanced the factual grounding of generated responses. Similarly, Khan et al. [10] focused on embedding models for processing PDF documents. Their experiments with OpenAI's "text-embedding-ada-002" model demonstrated a 35% increase in retrieval precision and a 28% reduction in latency for document-heavy fields like engineering. These findings underscore the critical role of embedding quality in ensuring accurate and efficient information retrieval in RAG systems. The technical architecture of vector databases has evolved significantly to complement embedding models. Han et al. [11] conducted a detailed survey of indexing techniques for vector databases, categorizing approaches into hash-based, tree-based, graph-based, and quantization-based methods. Their findings showed that graph-based approaches, particularly Hierarchical Navigable Small Worlds (HNSW), achieved superior performance with 99.3% recall at k=10 and search times of less than 1ms for datasets exceeding 10 million vectors.

## 2.3. Applications of RAG in Knowledge-Intensive Domains

In engineering, RAG systems have supported complex tasks. Buehler [12] demonstrated that integrating RAG with ontological knowledge graphs enhances accuracy and detail in material design. The study introduced MechGPT, a fine-tuned model for materials mechanics, which, combined with ontological graphs, improved concept discernment and interrelations. This approach boosted response accuracy and relevance while providing interpretable, information-rich graph structures. These findings highlight the value of domain-specific embedding models in improving factual accuracy and relevance.

Siddharth and Luo [13] demonstrated the utility of fine-tuning LLMs for de-sign engineering tasks by employing relation extraction techniques. Their dataset-specific approach achieved a remarkable 99.7% accuracy in token classification, underscoring the value of tailored datasets and task-specific embedding methods. Similarly, Machado [14] applied Retrieval-Augmented Generation (RAG) in industrial maintenance, revealing that prompt specificity plays a crucial role in retrieval accuracy. The study developed a cognitive assistant for industrial maintenance at STMicroelectronics, utilizing the RAG methodology to combine generative language models with the retrieval of specific information, thereby enhancing the system's response capability. These findings highlight the necessity of domain expertise in designing effective prompts and retrieval mechanisms to ensure accurate and contextually relevant responses.

## 3. METHODOLOGY

This section discusses the models incorporated into the RAG pipeline, the justification for their selection, the hyperparameters employed, the chunking strategies implemented, the libraries utilized, the retrieval mechanisms applied, and other pertinent elements of the system's design

### 3.1. Dataset

The dataset selected focuses specifically on ABB circuit breakers, particularly the SACE Emax 2 low-voltage air circuit breakers. This dataset was chosen due to its widespread application in various settings and facilities, including datacentres, hospitals, and more. The data is readily accessible on the ABB website and is the same information that electrical engineers rely on for the design and operation of these breakers, considering various characteristics and functionalities during the de-sign of electrical systems. The dataset provides detailed explanations of the assembly and disassembly processes of circuit breakers. However, the complete information was not included. Instead, a carefully selected subset of documents, focused on one series of breakers, was used to ensure a comprehensive yet efficient analysis.

Table 1 Dataset Size (in tokens).

| Document | Number of Tokens |
|---|---|
| Emax E1.2 Touch Control: Ground Fault Guide | 507 |
| Emax E1.2 Touch Control Instructions | 706 |
| Emax E1.2 DIP Control: Ground Fault Guide | 503 |
| ABB Circuit Breaker Ratings Explained | 2,367 |
| ABB Solutions for Datacentres | 4,290 |
| How to Disassemble Emax E1.2 Circuit Breaker | 8,738 |
| Emax E1.2 Circuit Breaker Details | 174,040 |
| Emax E1 Circuit Breaker Overview | 18,301 |
| Emax Circuit Breaker Technical Info | 19,139 |
| Emax2 Circuit Breaker Standards Guide (July 2024) | 253,830 |
| Emax Circuit Breaker Communication Features (Aug 2021) | 22,036 |
| Low Voltage Circuit Safety (Feb 2024) | 33,251 |

The 12 files outlined in Table 1, tokenized into 537,708 tokens using GPT4o, form an interconnected dataset. It would be impractical for an electrical engineering designer, responsible for ensuring breaker parameters are set correctly, to create an electrical design centered around this breaker without access to most of these documents. These documents help configure the settings accurately and are equally valuable for operations engineers tasked with maintaining such equipment. They provide insights into the designer's rationale and detailed guidance on the assembly and disassembly processes of the circuit breaker for ease of maintenance.

### 3.2. Chunking Methods

To address the extensive number of characters and tokens within the documents, chunking strategies were employed using the Unstructured library's partition_pdf tool.[1] This library was chosen due to its support for tables and its ability to be used with minimal preprocessing, thereby expediting the development of RAG systems and reducing the time required for data cleaning and preparation. Three distinct chunking methods were utilized to determine the most effective strategy:

Basic Chunking Strategy: This method segments the documents without considering section boundaries, page boundaries, or content similarity. Using parameters such as max_characters=1000 and new_after_n_chars=800, it limits chunk size for precision and token efficiency. Smaller elements are grouped with combine_text_under_n_chars=500.

Paragraph-per-Page Chunking: Although not a standard chunking strategy offered by the Unstructured library, this method was implemented by setting multipage_sections=False to enforce page boundaries. With max_characters=1500 and new_after_n_chars=1200, it balances chunk size and paragraph segmentation, grouping smaller elements using combine_text_under_n_chars=700.

By-Title Chunking: This method creates a new chunk whenever a title is detected. Parameters like max_characters=4000 and new_after_n_chars=3800 support extended context retention, while combine_text_under_n_chars=2000 ensures small-er sections are integrated logically.
These chunking methods were evaluated to identify the most suitable approach for optimizing retrieval and maintaining semantic coherence within the dataset.

### 3.3. Raw Text Summarization

The text that was chunked was summarized using GPT4o-mini for easier retrieval of the raw text. A temperature of 0 was used for the summary creation to maintain consistency and reproducibility.

### 3.4. Vector Database

The study used Chroma, a free, serverless, and open-source vector store li-censed under Apache 2.0, chosen for its robust integration with embedding models and efficient handling of large datasets. Its serverless nature simplifies deployment and reduces overhead, while its open-source framework allows extensive customization. This flexibility enabled rapid experimentation with embedding models, optimizing performance and efficiency. [2]

### 3.5. Models and Embeddings

The RAG systems under evaluation incorporate three distinct pipelines, each utilizing a different combination of models and embedding methods tailored to the respective model's characteristics.

---

[1] https://docs.unstructured.io/open-source/core-functionality/partitioning#partition
[2] http://trychroma.com.

Cohere Model Pipeline: The first model integrated is Cohere's "command-xlarge-nightly"[3], chosen for its optimization toward performance tasks as highlighted in the Cohere website release notes. This model prioritizes performance over response speed, aligning with the study's objectives of evaluating system effectiveness. It features an input context window of 128,000 tokens and a maximum output token size of 4,000 tokens. The embedding model utilized for this pipeline's vector store is "em-bed-english-v3.0"[4] from Cohere. The model was limited to an input context token size of 4000, and an output token size of 4000 to ensure accuracy and consistency withing the evaluation process.

OpenAI GPT4o Pipeline: The second pipeline employs OpenAI's GPT4o[5], the latest release known for its superior performance and efficiency. It features an input context window of 128,000 tokens. The embedding model used in this pipeline is "text-embedding-ada-002"[6]. The model was limited to an input context token size of 4000, and an output token size of 4000 to ensure accuracy and consistency within the evaluation process.

Anthropic Claude Pipeline: The third model integrated into the RAG systems is Anthropic's "Claude-3-5-sonnet-20240620."[7] Although not the newest model at the time of writing, this version was the most up to date during the initial phases of the study and was retained for consistency. This model offers a context window of 200,000 tokens. Since Anthropic does not provide an embedding model, Voyage AI's "voyage-3"[8] embedding model was adopted. The embedding model was selected based on the recommendation provided on the official Anthropic website[9]. The model was limited to an input context token size of 4000, and an output token size of 4000 to ensure consistency within the evaluation process.

This context length was chosen to mitigate issues observed in previous configurations, where a shorter context of 2,000 tokens resulted in incomplete responses and insufficiently accurate outputs. By increasing the context length, the model's ability to generate complete and contextually aligned answers was significantly enhanced.

These pipelines provide a diverse and robust framework for evaluating the performance of RAG systems across varying configurations, allowing for a comprehensive analysis of their applicability and effectiveness in domain-specific tasks.

### 3.6. Document Retrieval and Ranking Mechanism

To enable document retrieval and ranking, the embedding functions were integrated with the Chroma Vector Store. Queries are transformed into semantic vectors using the embedding models, enabling similarity-based retrieval with document embeddings. The retrieval prioritizes recall, fetching up to 2k documents (default k=10). Retrieved documents are re-embedded to improve precision, and ranking is performed using cosine similarity calculated with the sklearn.metrics.pairwise.cosine_similarity function.[10]

---

[3] https://docs.cohere.com/v2/docs/models
[4] https://docs.cohere.com/v2/docs/embeddings
[5] https://platform.openai.com/docs/models/gp#gpt-4o
[6] https://platform.openai.com/docs/guides/embeddings/embedding-models
[7] https://docs.anthropic.com/en/docs/about-claude/models#model-comparison-table
[8] https://docs.voyageai.com/docs/embeddings
[9] https://docs.anthropic.com/en/docs/build-with-claude/embeddings
[10] https://scikit-learn.org/dev/modules/generated/sklearn.metrics.pairwise.cosine_similarity.html.

### 3.7. Hyperparameters for Top K docs

The retrieval and preparation of relevant documents to construct a query-specific context is done using the retrieve_with_ranking method, the top 10 most relevant documents are identified through a ranking process. To ensure compatibility with downstream processing constraints, such as token limits for language models, the context is truncated if its word count exceeds the specified maximum (max_context_tokens). This truncation retains only the initial portion of the context, adhering to the token limit while preserving relevance.

### 3.8. Temperature setting of the RAG Models Generation

A temperature setting of 0 was used for response generation to maintain consistency and reproducibility. This setting minimizes randomness, ensuring that the model produces identical responses for identical queries, a critical requirement in engineering contexts to prevent the dissemination of inconsistent or erroneous information.

### 3.9. Performance Metrics

RAGAS[11] was utilized to evaluate the performance of the RAG pipelines under examination. A question-answer set was carefully constructed by reviewing the documents to ensure the relevance and accuracy of the questions and answers. The dataset comprises a total of 31 questions derived from various documents. The selected metrics for RAGAS evaluation are as follows:

Faithfulness: This metric evaluates how factually consistent the generated answer is with the provided context. It is determined based on the alignment between the answer and the retrieved context, with a score ranging from 0 to 1, where a score closer to 1 indicates higher factual accuracy.

$$\text{Faithfulness score} = \frac{|\text{Number of claims in the generated answer that can be inferred from given context}|}{|\text{Total number of claims in the generated answer}|}$$

Context recall: This metric evaluates the proportion of relevant information retrieved compared to the reference context. Non-LLM-based context recall, used in this evaluation, relies on two inputs (retrieved and reference contexts) and is calculated using the following formula:

$$\text{context recall} = \frac{|\text{Number of Retrieved Contexts That Are Relevant}|}{|\text{Total number of Contexts in the reference}|}$$

Context precision: This metric measures the percentage of relevant information in retrieved chunks, calculated using the average precision@k. Precision@k is the ratio of relevant chunks at rank k to the total chunks at rank k. The formulas for k and context precision are shown below:

$$\text{Context Precision@K} = \frac{\sum_{k=1}^{K}(\text{Precision@k} \times v_k)}{\text{Total number of relevant items in the top } K \text{ results}}$$

$$\text{Precision@k} = \frac{\text{true positives@k}}{(\text{true positives@k} + \text{false positives@k})}$$

Answer relevance: This metric assesses the relevance of the generated answer to the question, considering the retrieved context. It calculates the mean cosine similarity between the original user input and a set of artificially generated questions, which are reverse engineered from the response. The equation for it is below:

$$\text{answer relevancy} = \frac{1}{N}\sum_{i=1}^{N} \cos(E_{g_i}, E_o)$$

---

[11] https://docs.ragas.io/en/stable/.

N is 3 by Default, Egi is the embedding of the generated question, and Eo is the embedding of the original question.

## 4. EXPERIMENTAL RESULTS AND DISCUSSION

The experiments were conducted using Google Colab as the computational environment. The system specifications included an Intel Xeon CPU with 2 virtual CPUs (vCPUs), 13 GB of RAM, and 108 GB of disk space. Multiple APIs were utilized during the experiments, including the OpenAI API, VoyageAI API, Claude API, and Cohere API, to enable seamless integration and execution of the different RAG pipelines.

Comprehensive analyses and experiments were conducted on each of the models, utilizing their specific embedding configurations and evaluating them across the distinct chunking methods described earlier. To standardize the evaluation process, a consistent system prompt was provided to each model. The prompt instructed:

"You are an electrical engineer tasked with analyzing technical data and providing engineering insights. You will be given a mixture of text, tables, and image(s), which usually contain schematics, technical diagrams, or data charts. Use this information to provide engineering advice related to the user question. If you do not know, respond with 'I don't know'."

This prompt was leveraged uniformly across the three models—Cohere, Claude, and GPT4o—to ensure comparability and consistency in the evaluation process.

The parameters configured for the Cohere RAG pipeline were optimized to ensure the efficient processing and retrieval of contextually relevant information. The performance of the Cohere RAG pipeline was evaluated across the three chunking methods described earlier, beginning with the basic chunking strategy. Results for the Cohere RAG pipeline using these chunking methods are presented in Table 2, highlighting the system's performance metrics as assessed by RAGAS.

Table 2. Cohere Chunking Methods RAGAS Evaluation

| Model | Chunking Method | Context Precision | Faithfulness | Answer Relevancy | Context Recall |
|---|---|---|---|---|---|
| **Cohere** | Basic | 0.9697 | 0.6711 | 0.7136 | 0.5697 |
| | Paragraph per Page | 1.0000 | 0.7778 | **0.7562** | **0.7705** |
| | By Title | **1.0000** | **0.7965** | 0.7263 | 0.5809 |

For Context Precision, both the "Paragraph per Page" and "By Title" methods achieved the highest possible score (1.0000), demonstrating excellent alignment with the query context. In terms of Faithfulness, the "By Title" method led with a score of 0.7965, reflecting superior factual accuracy. However, the "Paragraph per Page" method outperformed others in Answer Relevancy, scoring 0.7562, showcasing its ability to provide answers more relevant to the context. Furthermore, the "Paragraph per Page" method excelled in Context Recall, achieving the highest score of 0.7705, indicating its strength in retrieving a broader range of relevant information.

Overall, the "Paragraph per Page" method is identified as the best-performing approach. While the "By Title" method slightly outpaces it in Faithfulness (0.7965 vs. 0.7778), the "Paragraph per Page" method's superior performance in Answer Relevancy and significantly higher Context Recall make it the more practical choice.

The second model integrated into the RAG pipeline is OpenAI's GPT4o. For consistency and to facilitate a balanced comparison across models, the same token length parameters were applied as those used for the Cohere model. This configuration ensures that the model has sufficient context to generate comprehensive and relevant answers while maintaining uniformity across the RAG pipelines for evaluation purposes.

The performance of the OpenAI GPT4o RAG pipeline was evaluated for each of the chunking methods outlined earlier. Detailed results are presented in Table 3:

Table 3. GPT4o Chunking Methods RAGAS Evaluation

| Model | Chunking Method | Context Precision | Faithfulness | Answer Relevancy | Context Recall |
|---|---|---|---|---|---|
| **GPT4o** | Basic | 0.9677 | 0.6820 | 0.7385 | **0.6984** |
| | Paragraph per Page | 0.9355 | 0.6625 | 0.7102 | 0.6491 |
| | By Title | **1.0000** | **0.7741** | **0.8021** | 0.6073 |

For Context Precision, the "By Title" method (Method 3) achieved the highest possible score (1.0000), showcasing exceptional alignment of retrieved information with the query context. Regarding Faithfulness, Method 3 also excelled with the highest score (0.7741), indicating superior factual accuracy. Additionally, Method 3 led in Answer Relevancy with a score of 0.8021, demonstrating its ability to provide highly contextually relevant answers. However, in terms of Context Recall, the "Basic" method (Method 1) performed best with a score of 0.6984, retrieving a broader range of relevant information.

Overall, Chunking Method 3 ("By Title") is identified as the most effective approach. Despite its lower Context Recall (0.6073), its significantly higher scores in Context Precision, Faithfulness, and Answer Relevancy establish it as the optimal choice for accuracy- and relevance-focused applications. This makes Method 3 particularly well-suited for precision-critical use cases, such as ABB breaker documentation and other high-stakes scenarios.

For the final model evaluated, Claude, identical parameters were utilized as those established for GPT4o and Cohere. Specifically, the token length for context and response was maintained, and the temperature setting remained at 0 to ensure factual consistency. Performance results for the Claude model are presented in Table 4:

Table 4. Claude Chunking Methods RAGAS Evaluation

| Model | Chunking Method | Context Precision | Faithfulness | Answer Relevancy | Context Recall |
|---|---|---|---|---|---|
| **Claude** | Basic | 1.0000 | 0.7930 | 0.7375 | **0.7216** |
| | Paragraph per Page | **1.0000** | **0.8556** | 0.7410 | 0.6913 |
| | By Title | 1.0000 | 0.8342 | **0.7732** | 0.6763 |

For Context Precision, all three methods achieved perfect scores (1.0000), indicating flawless alignment of retrieved information with the query context. In terms of Faithfulness, Method 2 ("Paragraph per Page") emerged as the top performer with the highest score (0.8556), reflecting exceptional factual accuracy. Answer Relevancy was dominated by Method 3, which scored 0.7732, highlighting its ability to deliver highly contextually relevant answers. For Context Recall, Method 1 achieved the highest score (0.7216), demonstrating its strength in retrieving a broader range of relevant information.

Overall, Chunking Method 2 ("Paragraph per Page") stands out as the most effective approach. While Method 3 leads in Answer Relevancy (0.7732), Method 2 provides a superior balance, combining the highest Faithfulness (0.8556) with competitive Context Recall (0.6913). This balanced performance makes Method 2 particularly well-suited for applications prioritizing factual correctness and precision, such as technical documentation retrieval and domain-specific queries.

The evaluation identified distinct best and worst performers across metrics. For Context Precision, Cohere's and Claude's "By Title" and "Paragraph per Page" methods scored perfectly (1.0000), while GPT4o's "Paragraph per Page" method scored lowest (0.9355). In Faithfulness, Claude's "Paragraph per Page" method was the most accurate (0.8556), while GPT4o's scored lowest (0.6625). For Answer Relevancy, GPT4o's "By Title" method led (0.8021), while Cohere's "Basic" method scored lowest (0.7136). In Context Recall, Cohere's "Paragraph per Page" method excelled (0.7705), while its "Basic" method scored lowest (0.5697). These results highlight trade-offs across evaluation criteria.

In the final ranking, the best overall method was Claude's "Paragraph per Page," achieving the highest Faithfulness (0.8556) and excellent performance across other metrics, making it the most accurate and reliable approach. The runner-up was Cohere's "Paragraph per Page," offering a well-balanced performance with the highest Context Recall (0.7705), making it ideal for broader information retrieval. The worst overall method was Cohere's "Basic," with the lowest scores in Context Recall (0.5697) and Answer Relevancy (0.7136), rendering it the least effective option.

**Qualitative Analysis:**

Diving into the answers that are generated by each model and chunking method for a set of questions carefully selected to see how each model coupled with the different chunking methods approached the questions and answers them.

**Question 1:** "How can the terminating resistance (Rterm) be activated on the Ekip Com Modbus RTU module, and what is the role of dip-switches in this process?".

This question is valuable for the qualitative analysis of RAG answers as it tests the system's ability to retrieve domain-specific, technically accurate, and contextually relevant information. It evaluates the model's precision in addressing specialized topics, such as activating terminating resistance (Rterm) and configuring dip-switches in Modbus RTU modules, while assessing answer relevancy and technical faithful-ness. Additionally, the practical implications of the question align with end-user needs, making it an effective metric for evaluating the RAG system's real-world applicability and utility. The answer of each chunking method can be found in short below:

Cohere Basic Chunking method: "The Ekip Com Modbus RTU module does not mention a specific terminating resistance feature, but the Ekip Com Profibus DP module includes a 220 Ω termination resistor that can be activated using dip-switches. These switches allow manual configuration of the resistor for proper bus termination, ensuring signal integrity on the RS-485 bus. Dip-switches are compact, manual switches used for configuring device parameters directly on circuit boards."

Cohere Paragraph per Page Chunking method: "The process of activating the terminating resistance (Rterm) on the Ekip Com Modbus RTU module is not explicitly detailed in the provided text. However, the module's configuration typically involves setting parameters like baud rate and addressing via dip-switches. To activate Rterm, the relevant dip-switch(es) would need to be toggled to the "On" position, as indicated in the module's manual. Dip-switches provide a physical means for adjusting settings, including termination, allowing customization for communication requirements. For precise instructions, refer to the "Technical Application Handbook - vol. 9" or the manufacturer's documentation."

Cohere By Title Chunking method: "The Ekip Com Modbus RTU module does not explicitly feature a "terminating resistance (Rterm)" but supports termination resistors for the RS-485 bus, commonly configured using dip-switches or jumpers. By setting these switches to the appropriate positions, users can enable a 120 Ω termination resistor, as mentioned in the documentation. Dip-switches provide a straightforward way to configure settings, including termination, without software intervention. For exact configurations, refer to the module's user manual. This approach ensures proper bus termination and signal integrity in RS-485 communication systems."

GPT4o Basic Chunking method: "To activate the terminating resistance (Rterm) on the Ekip Com Modbus RTU module, use the module's dip-switches. The document specifies a 390 Ω pull-up or pull-down resistor for bus termination, adjustable via these switches. Dip-switches enable or disable the termination resistor, ensuring proper signal integrity on the RS-485 serial bus, particularly for long cables or high-speed communication. This configuration minimizes signal reflections and reduces communication errors.

GPT4o Paragraph per Page Chunking method: "To activate the terminating resistance (Rterm) on the Ekip Com Modbus RTU module, use its dip-switches. These manual switches enable or disable the terminating resistor, essential for maintaining signal integrity in a Modbus RTU network. The resistor matches the transmission line's impedance, reducing signal reflections and preventing communication errors. It is typically activated when the module is at the end of the communication line. Refer to ABB documentation for precise dip-switch settings and configuration instructions.

GPT4o By Title Chunking method: "To activate the terminating resistance (Rterm) on the Ekip Com Modbus RTU module, use the dip-switches to con-figure a 120 Ω termination resistor for the RS-485 bus. This resistor ensures proper signal integrity and minimizes reflections, especially when the module is at the end of the bus. Termination resistors should only be placed at the bus ends, not on branch lines. Refer to ABB documentation, such as 1SDH001140r0001, for detailed dip-switch configuration instructions."

Claude Basic Chunking method: "The terminating resistance (Rterm) on the Ekip Com Modbus RTU module can be activated using dip-switches, which enable a 390 Ω pull-up or pull-down termination resistor. This resistor ensures proper termination of the RS-485 serial bus, reducing signal reflections and maintaining reliable Modbus RTU communication. Terminating resistors should only be activated at the network's ends to match the cable's characteristic impedance. Dip-switches offer a flexible and convenient method for enabling or disabling the resistor without requiring hardware modifications. Proper configuration is crucial to avoid communication issues, as detailed in ABB's documentation.

Claude Paragraph per Page Chunking method:" The terminating resistance (Rterm) on the Ekip Com Modbus RTU module can be activated using dip-switches located on the module. These switches provide a simple physical means to enable or disable the terminating resistor, typically set to 120 Ω, without requiring software changes. Proper termination ensures signal integrity by preventing reflections on the RS-485 bus, critical for reliable communication in Modbus RTU networks. Activate the resistor if the module is at the bus's end, and deactivate it if positioned in the middle. Refer to the module's documentation for specific switch configurations."

Claude By Title Chunking method: "The terminating resistance (Rterm) on the Ekip Com Modbus RTU module can be activated using dip-switches, which con-figure the standard 120 Ω resistor for RS-485 networks. To enable it, set the appropriate dip-switch to the ON position, typically for modules at the ends of the communication bus, ensuring proper signal integrity and preventing reflections. Devices in the middle of the bus should not activate their resistors. Dip-switches provide a straight-forward, hardware-based method to manage termination without requiring software changes. Proper configuration is critical for reliable communication, especially in longer or noisy RS-485 networks. Additional resistors (220 Ω and 390 Ω) may be required as part of the overall network design.

**Reference answer:** "Terminating resistance On the Ekip Com Modbus RTU module it is possible to activate the terminating resistance Rterm= 120 Ω. To enable the Rterm, the corresponding dip-switches 1 and 2 must be positioned to ON (see previous figure). With the Ekip Com Modbus RTU modules, the dip-switches 3 and 4 of the Rpol (polarization resistance), are not used."

Comparing the responses generated by Cohere's three chunking methods, The By Title chunking method provides the most accurate answer, identifying the 120 Ω terminating resistance. However, it does not specify the positions of the dip switches, which are critical in this context. The Basic chunking method produces an incorrect answer, mentioning a 220 Ω resistance, which could result in confusion and reduced trust in the model's reliability for further inquiries by engineers. The Para-graph per Page chunking method fails to mention the resistor, an essential component in this scenario. Notably, all chunking methods accurately describe the function of the dip switches.

For GPT4o, the By Title chunking method again provides the most accurate response by correctly identifying the 120 Ω resistance but similarly fails to specify the dip switch positions. The Paragraph per Page chunking method accurately explains the use of dip switches but omits any mention of the resistance. The Basic chunking method produces an incorrect response, referencing a 390 Ω resistance.

Claude's responses indicate that the best results are obtained from both the By Title and Paragraph per Page chunking methods, aligning with the quantitative analysis. The Basic chunking method, similar to GPT4o, erroneously mentions a 390 Ω resistance. Overall, the Paragraph per Page chunking method in Claude provides the most comprehensive and useful information without introducing misleading details.

However, all models fail to correctly identify the dip switch positions, which are essential for the query. This failure might stem from the question not explicitly emphasizing the need to identify dip switch positions.

**Question 2:** "What are the recommended busbar dimensions and configurations for the E1.2 circuit breaker with a current rating (Iu) of 630A for both horizontal and vertical terminals?". This question is important as it addresses the technical specifications essential for ensuring the safe and efficient operation of the E1.2 circuit breaker at a 630A current rating. Proper busbar dimensions and configurations are critical for thermal management, minimizing electrical resistance, and ensuring mechanical compatibility, directly impacting system reliability and compliance with industry standards. Additionally, understanding the configurations for both horizontal and vertical terminals provide flexibility in installation, catering to diverse application requirements. The answers can be found below:

All models and chunking methods "For the E1.2 circuit breaker (Iu 630A), the recommended busbar dimensions are 2x40x5 mm for both horizontal and vertical terminals, ensuring proper performance and safety under IEC 60947 standards."

**Reference answer:** "For the E1.2 circuit breaker with a current rating (Iu) of 630A, the recommended busbar dimensions are 2x40x5 mm for both horizontal and vertical terminals, as specified by the IEC 60947 standard. These dimensions ensure proper connectivity and reliable performance for the circuit breaker."

The analysis of this question is straightforward, all answers and chunking methods got the correct answers. There were some differences between each model and chunking method, however they were not critical to be explored further.

**Question 3:** "What are the available ground fault pickup and time delay set-tings for the Emax E1.2 circuit breaker with Ekip DIP LSI/LSIG, and which standard do they comply with?". This question is important because it addresses the specific safety and performance settings of the Emax E1.2 circuit breaker with Ekip DIP LSI/LSIG, focusing on ground fault protection. Understanding the available pickup and time delay settings is crucial for ensuring compliance with electrical safety standards, such as IEC 60947. These settings directly influence the circuit breaker's ability to detect and respond to ground faults, which are critical for preventing damage, enhancing system reliability, and ensuring personnel safety in electrical systems. The answer of each chunking method can be found in short below:

Cohere Basic Chunking method: "The Emax E1.2 circuit breaker with Ekip DIP LSI/LSIG has a ground fault pickup setting of 50,000 Amperes (-16% to 0% tolerance) and supports current ratings of 1200, 250, 400, and 800 Amperes. Operating at 50-60 Hz, it functions in temperatures from -25°C to +70°C and altitudes up to 2000 meters. While time delay settings and specific standards are not mentioned, its detailed specifications ensure reliable performance and compliance with safety requirements."

Cohere Paragraph per Page Chunking method: "The ground fault protection settings for the Emax E1.2 circuit breaker with Ekip DIP LSI/LSIG are detailed in TCC 9AKK108468A2274, which likely specifies pickup and time delay values. These settings adjust for temperatures above +40°C. The breaker complies with IEC standards for performance and electromagnetic compatibility."

Cohere By Title Chunking method: "The ground fault pickup and time de-lay settings for the Emax E1.2 circuit breaker with Ekip DIP LSI/LSIG are not specified in the text but are likely detailed in TCC 9AKK108468A2274. While the specific standard compliance is not stated, the breaker supports temperatures up to +70°C, altitudes up to 2000 meters, and an interrupting rating of 50,000 Amperes (-16% to 0% tolerance), indicating compliance with industry standards like IEC or UL. Refer to TCC 9AKK108468A2274 and technical documentation for exact settings and standards."

GPT4o Basic Chunking method: "The Emax E1.2 circuit breaker with Ekip DIP LSI/LSIG has a ground fault pickup setting of 50,000 Amperes (-16% to 0% tolerance), but specific time delay settings are not provided. For detailed ground fault settings, refer to document TCC 9AKK108468A2274. The breaker is generally designed to comply with IEC 60947-2, but specific compliance details should be confirmed through the manufacturer's documentation."

GPT4o Paragraph per Page Chunking method: "The ground fault pickup and time delay settings for the Emax E1.2 circuit breaker with Ekip DIP LSI/LSIG are not detailed in the text but are referenced in TCC 9AKK108468A2274. The breaker complies with IEC 61557-12, ensuring 1% precision for electrical measurement performance. For exact settings, consult the mentioned document."

GPT4o By Title Chunking method: "I don't know."

Claude Basic Chunking method: "The ground fault pickup settings for the Emax E1.2 circuit breaker with Ekip DIP LSI/LSIG range from 0.1 to 1.0 in increments of 0.001, with a maximum of 1,200A. Time delay settings are available at 0.1, 0.2, and 0.4 seconds, adhering to IEEE/ANSI standards. While these settings are inferred from general low-voltage power circuit breaker information, specific details for this model should be confirmed through ABB's documentation or technical support."

Claude Paragraph per Page Chunking method:" The Emax E1.2 circuit breaker with Ekip DIP LSIG includes ground fault protection (G protection), with settings referenced in document TCC 9AKK108468A2274. It complies with IEC 60255-151 for protection functions and IEC 61557-12 for 1% precision. Ground fault protection requires a current sensor connected via a two-pole shielded cable (e.g., BELDEN 9841, max 15 meters). Thresholds and protections can be configured using the Ekip T&P module and Ekip Connect software. For exact pickup and time delay settings, refer to the TCC document or ABB support"

Claude By Title Chunking method: "The provided information does not include specific ground fault pickup and time delay settings for the Emax E1.2 circuit breaker with Ekip DIP LSI/LSIG. However, these settings are likely detailed in the referenced document, TCC 9AKK108468A2274. Compliance with specific standards is not explicitly mentioned. For precise details, consult the TCC document or the manufacturer's product manual."

**Reference answer:** "The Emax E1.2 circuit breaker with Ekip DIP LSI/LSIG offers ground fault pickup settings at multiples of the rated current (In), including 0.1, 0.2, 0.3, 0.4, 0.6, 0.8, and 1. The available time delay settings are 0.1, 0.2, and 0.4 seconds. These configurations comply with the IEEE/ANSI C37.17 standard."

Most of the answers analyzed, while not incorrect, fail to effectively address the question. The responses provided by each model vary significantly based on the chunking method used, highlighting the critical role of chunking strategies and their parameters in delivering meaningful context to the model.

For Cohere, both the Basic and By Title chunking methods reference the ground fault pickup current of 50,000 amperes and include a citation of the technical manual as a source for further details. In contrast, the Paragraph per Page method references only the technical manual, omitting any information about the ground fault pickup current. Despite these efforts, none of the Cohere methods produce an answer that aligns closely with the reference answer.

In the case of GPT4o, the Basic chunking method similarly references the ground fault pickup current and the technical manual, whereas the Paragraph per Page method mentions only the manual. Interestingly, the By Title method simply states "I don't know," demonstrating a surprising limitation. This discrepancy suggests that, although the By Title method achieves a higher rating overall, it still has significant gaps requiring further investigation.

For Claude, the Basic chunking method performs relatively well, correctly identifying the time delay settings, a key part of the question. However, the Para-graph per Page and By Title methods fail to provide any substantial information beyond referencing the technical manual.

Overall, the analysis emphasizes that chunking methods have a profound impact on a model's ability to retrieve and generate relevant information, with significant variations in performance based on the approach employed. Further refinement and exploration of chunking strategies are necessary to improve alignment with reference answers.

**Addressing Data Gaps in RAG Responses**

The presence of "I don't know" responses in the Retrieval-Augmented Generation (RAG) system highlights existing limitations in retrieval effectiveness, embedding accuracy, and chunking strategies. These responses indicate potential gaps where the system fails to retrieve or interpret the necessary information, thereby impacting its reliability in engineering applications. Several factors contribute to these gaps, including incomplete context retrieval, suboptimal chunk segmentation, query-document mismatches, and constraints imposed by the model's context window. Additionally, the lack of fine-tuning on ABB-specific documentation reduces the system's ability to generate precise responses for domain-specific queries.

To address these limitations, several mitigation strategies can be implemented:

**Optimized Chunking Strategies**

A hybrid chunking approach that combines By Title and Paragraph-per-Page methods can enhance the system's retrieval efficiency by balancing contextual precision and recall. This approach ensures that segments retain semantic coherence while improving retrieval alignment.

**Enhanced Retrieval Mechanisms**

The integration of reranking models, such as Cohere Rerank or OpenAI Rerank, can prioritize the most relevant retrieved documents. Additionally, multi-stage retrieval techniques can refine initial broad retrievals through query expansion, ensuring that the system identifies the most contextually relevant sources.

**Iterative Query Refinement and Multi-Step RAG**

Inspired by i-MedRAG methodologies in the medical field, incorporating iterative query refinement can allow the system to generate follow-up queries when uncertain responses are detected. Additionally, a multi-turn retrieval pipeline can iteratively refine query inputs to ensure completeness before generating responses.

## 5. CONCLUSIONS, FUTURE WORK AND ETHICAL ISSUES

This study highlights critical limitations in using RAG for engineering, where approximately 80% efficiency is inadequate for a field intolerant of inaccuracies. Such errors raise ethical concerns about the reliability and safety of AI-generated recommendations in high-stakes contexts, emphasizing the need for rigorous validation and accountability.

Furthermore, the ethical issue of transparency is crucial; engineers must be able to understand and trust the decision-making processes of RAG systems. Without clear explainability, reliance on these models could lead to errors that compromise both safety and system performance. Additionally, incorporating data from multiple circuit breakers introduces the risk of compounding inaccuracies, highlighting the need for careful curation and verification of training data to ensure compliance with industry standards and avoid misleading outputs.

Despite these challenges, the rapid advancements in this area, including innovations in models, chunking methods, embeddings, and vector databases, provide substantial opportunities for improvement, with new developments emerging frequently. Future research in this domain could explore various directions, such as incorporating knowledge graphs, leveraging multimodal systems, experimenting with alternative models or upgraded versions of those utilized in this study, and investigating additional methodologies.

A key limitation of this system was its focus on a single circuit breaker, highlighting challenges in scaling. Expanding to multiple circuit breakers could increase complexity and the risk of inaccuracies, potentially undermining trust and usability. Future research must address these issues to enable ethical and reliable deployment of RAG systems in engineering. By overcoming these challenges and leveraging advancements, RAG systems can better meet the demands of engineering applications while upholding accuracy, transparency, and reliability.

## 5.1. Enhancing RAG with Knowledge Graphs and Multimodal Data

Integrating knowledge graphs (KGs) and multimodal data into RAG systems can address retrieval gaps and improve response accuracy in technical domains such as electrical engineering. These enhancements refine contextual alignment, factual consistency, and retrieval precision.

**Knowledge Graph Integration**

Knowledge graph's structure relationships between entities, improving retrieval accuracy and fact verification. In the context of ABB circuit breakers, a KG could model relationships between:

Breaker models (e.g., Emax E1.2, Tmax XT),
Technical specifications (e.g., current ratings, trip settings),
Safety standards (e.g., IEC 60947, IEEE C37.17).

Integrating a KG-augmented retriever allows for:

1. Query Expansion – Enriching queries by linking related entities (e.g., associating "ground fault settings" with "Ekip DIP LSIG").
2. Fact Verification – Cross-checking generated responses against structured data.
3. Improved Ranking – Prioritizing documents based on KG-driven semantic relevance.

**Multimodal Data Integration**

Technical queries often require non-textual references, such as schematics, tables, and scanned documents. A multimodal RAG pipeline could incorporate:

1. Image-Based Retrieval – Using Vision Transformers (ViTs) to interpret wiring diagrams and schematics.
2. Table Parsing – Employing TAPAS or DeepTable for structured extraction of electrical parameters.

**Implementation Strategy**

Future enhancements should focus on:

1. KG-Augmented Retrieval, combining vector search with graph-based reasoning.
2. Hybrid Text-Image Retrieval, leveraging contrastive learning for joint multimodal processing.
3. Ontology-Based Query Expansion, using structured knowledge to enhance response precision.

Integrating knowledge graphs and multimodal retrieval can improve RAG's performance in engineering applications, ensuring greater contextual relevance, factual accuracy, and compliance with industry standards.

AUTHORS

**Salahuddin Alawadhi** Received his Bachelor's degree in engineering Currently, he is pursuing his Master of Science in Artificial Intelligence from the University of Leeds. His research interests include Retrieval-Augmented Generation (RAG), knowledge graphs, and the application of AI in electrical systems and renewable energy.

Dr. **Noorhan Abbas** received her Bachelor of Science in Computer Science and Business Administration from the American University in Cairo. She then earned her Master of Science in Artificial Intelligence and Text Analytics from the University of Leeds. Dr. Abbas completed her PhD in Technology and Policing at Lancaster University. Currently, she is a Research and Teaching Fellow at the University of Leeds, where she leads modules in Programming for Data Science and Data Mining and Text Analytics as part of the online MSc in Artificial Intelligence program. Her research interests include Artificial Intelligence, Natural Language Processing, Large Language Models, Educational Chatbots, Data Science, and Machine Learning.